%% file: main.tex
\pdfoutput=1
\documentclass[11pt]{article}
\usepackage[]{emnlp2021}
\usepackage{times}
\usepackage{latexsym}
\usepackage[T1]{fontenc}
\usepackage[utf8]{inputenc}
\usepackage{microtype}
\usepackage{amssymb}% http://ctan.org/pkg/amssymb
\usepackage{pifont}% http://ctan.org/pkg/pifont
\usepackage[ruled,vlined]{algorithm2e}
\include{pythonlisting}
\usepackage{listings}
\usepackage{dblfloatfix}
\usepackage{amsmath}
\usepackage{graphicx}
\usepackage{array,booktabs,ragged2e}
\usepackage{multirow}

\newcommand{\data}[0]{\textsc{Police Violence Frame Corpus}}
\newcommand{\adata}[0]{\textsc{PVFC}}

\newcolumntype{R}[1]{>{\RaggedLeft\arraybackslash}p{#1}}
\newcolumntype{L}[1]{>{\RaggedRight\arraybackslash}p{#1}}
\newcolumntype{M}[1]{>{\centering\arraybackslash}m{#1}}

\title{To Protect and To Serve? \\{A}nalyzing Entity-Centric Framing of Police Violence}

\author{Caleb Ziems \\
  School of Interactive Computing\\
  Georgia Institute of Technology \\
  \texttt{cziems@gatech.edu} \\\And
  Diyi Yang \\
  School of Interactive Computing\\
  Georgia Institute of Technology \\
  \texttt{dyang888@gatech.edu} \\}

\begin{document}
\maketitle
\begin{abstract}
Framing has significant but subtle effects on public opinion and policy. We propose an NLP framework to measure entity-centric frames. We use it to understand media coverage on police violence in the United States in a new \data{} of 82k news articles spanning 7k police killings. Our work  uncovers more than a dozen framing devices and reveals significant differences in the way liberal and conservative news sources frame both the \textsl{issue} of police violence and the \textsl{entities} involved. Conservative sources emphasize when the victim is armed or attacking an officer and are more likely to mention the victim's criminal record. Liberal sources focus more on the underlying systemic injustice, highlighting the victim's race and that they were unarmed. We discover temporary spikes in these injustice frames near high-profile shooting events, and finally, we show protest volume correlates with and precedes media framing decisions. \footnote{Data and code available at: \url{https://github.com/GT-SALT/framing-police-violence}
}
\end{abstract}

\section{Introduction}
The normative standard in American journalism is for the news to be neutral and objective, especially regarding politically charged events \cite{schudson2001objectivity}. Despite this expectation, journalists are unable to report on all of an event's details simultaneously. By choosing to include or exclude details, or by highlighting salient details in a particular order, journalists unavoidably induce a preferred interpretation among readers \cite{iyengar1990framing}. This selective presentation is called \textbf{framing} \cite{entman2007framing}. Framing influences the way people think by ``telling them what to think about'' \cite{entman2010media}. In this way, frames impact both public opinion \cite{chong2007framing,iyengar1990framing,mccombs2002agenda,price2005framing,rugg1941experiments,schuldt2011global} and policy decisions \cite{baumgartner2008decline,dardis2008media}. 

\begin{figure}
\centering
    \includegraphics[width=\columnwidth]{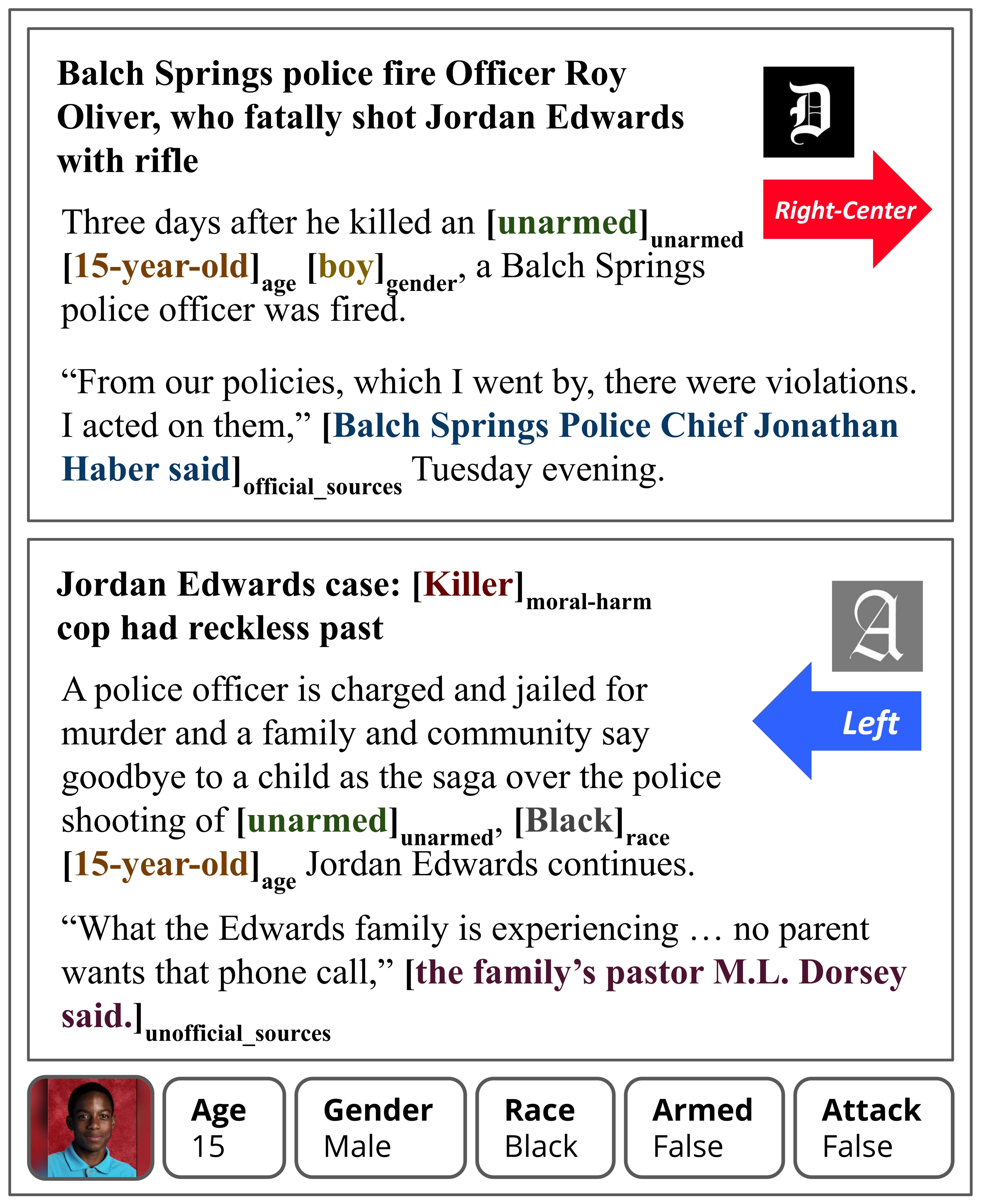}
    \caption{\small \textbf{Framing the murder of Jordan Edwards.} Our system automatically identifies key details or \textsl{frames} that shape a reader's understanding of the shooting. Importantly, we can distinguish the victim's attributes from the descriptions of the officer, like \textsl{killer} in ``killer cop.'' Only the left-leaning article uses this morally-weighted term, \textsl{killer}, and also takes care to mention the victim's \textsl{race}. While the left-leaning article highlights a quote from the Edwards' pastor, an \textsl{unofficial source}, the right-center article cites only \textsl{official sources} (namely the Chief of Police). Both mention the victim's \textsl{age} and \textsl{unarmed} status.
    }\label{fig:frame_example}
\end{figure}

Prior work has revealed an abundance of politically effective framing devices \cite{bryan2011motivating,gentzkow2010drives,price2005framing,rugg1941experiments,schuldt2011global}, some of which have been operationalized and measured at scale using methods from NLP \cite{card2015media,demszky2019analyzing,field2018framing,greene2009more,recasens2013linguistic,tsur2015frame}. While these works extensively cover \textsl{issue} frames in broad topics of political debate (e.g. \textsl{immigration}), they overlook a wide array of \textsl{entity} frames (how an individual is represented; e.g. a particular undocumented worker described as lazy), and these can have huge policy implications for target populations \cite{schneider1993social}.

In this paper, we introduce an NLP framework to understand \textsl{entity framing} and its relation to \textsl{issue framing} in political news. As a case study, we consider news coverage of police violence. Though we choose this domain for the stark contrast between two readily-discernible entities (police and victim), our framing measures can also be applied to other major social issues \cite{luodetecting,mendelsohn2021modeling}, and salient entities involved in these events, like protesters, politicians, migrants, etc. 

We make several novel contributions. First, we introduce the \data{} that contains 82k news articles on over 7k police shooting incidents. See Figure~\ref{fig:frame_example} for example articles with annotated frames. Next, we build a set of syntax-aware methods for extracting 15 issue and entity frames, implemented using entity co-reference and the syntactic dependency parse. Unlike bag-of-words methods (e.g. topic modeling) our \textbf{entity-centric methods} can distinguish between a white \textsl{man} and a white \textsl{car}. In this example, we identify race frames by scanning the attributive and predicative adjectives of any \textsc{victim} tokens. Such distinctions can be crucial, especially in a domain where \textsl{officer} aggression will have different ramifications than aggression from a \textsl{suspected criminal}. By exact string-matching, we can also extract, for the first time, differences in the \textbf{\textsl{order} that frames appear} within each document. 

We find that liberal sources discuss \textsl{race} and \textsl{systemic racism} much earlier, which can prime readers to interpret all other frames through the lens of injustice. Furthermore, we quantify and statistically confirm what smaller-scale content analyses in the social sciences have previously shown \cite{drakulich2020race,fridkin2017race,lawrence2000politics}, that conservative sources highlight law-and-order and focus on the victim's criminal record or their harm or resistance towards the officer, which could justify police conduct. Finally, we rigorously examine the broader interactions between media framing and offline events. We find that high-profile shootings are correlated with an increase in \textsl{systemic} and \textsl{racial} framing, and that increased protest activity \textsl{Granger-causes} or precedes media attention towards the victim's race and unarmed status.

\section{Related Work}
A large body of related work in NLP focuses on detecting stance, ideology, or political leaning \cite{baly2020we,bamman2015open,iyyer2014political,johnson2017ideological,preoctiuc2017beyond,luo2020desmog,stefanov2020predicting}. While we show a relationship between framing and political leaning, we argue that frames are often more subtle than overt expressions of stance, and cognitively more salient than other stylistic differences in the language of political actors, thus more challenging to be measured. 

\begin{figure*}[t]
\centering
    \includegraphics[width=0.85\textwidth]{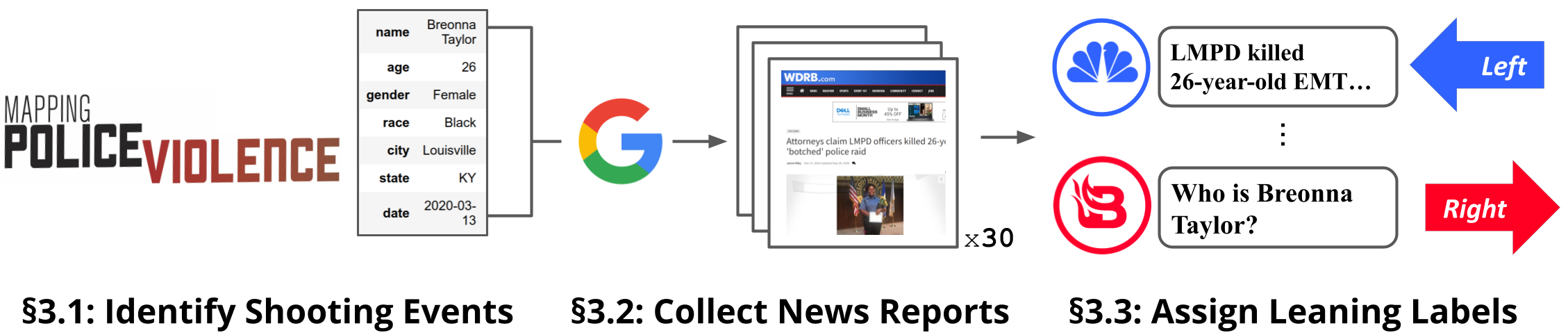}
    \caption{\small Construction process of \data{}.
    }
    \label{fig:pipeline}
\end{figure*}

Specifically, we distinguish between \textsl{issue} frames \cite{iyengar1990framing} and \textsl{entity} frames \cite{van2020doctor}. Entity frames are descriptions of individuals that can shape a reader’s ideas about a broader issue. The entity frames of interest here are the victim’s age, gender, race, criminality, mental illness, and attacking/fleeing/unarmed status. One related work found a \textsl{shooter identity} cluster in their topic model that contained broad descriptors like ``crazy'' \cite{demszky2019analyzing}. However, their bag-of-words method would not differentiate a crazy \textsl{shooter} from a crazy \textsl{situation}. To follow up, there is need for a \textsl{syntax-aware} analysis. 

In a systematic study of issue framing, \citet{card2015media} applied the Policy Frames Codebook of \citet{boydstun2013identifying} to build the Media Frames Corpus (MFC). They annotated spans of text from discussions on tobacco, same-sex marriage, and immigration policy with broad meta-topic framing labels like \textsl{health and safety}. \citet{field2018framing} built lexicons from the MFC annotations to classify issue frames in Russian news, and \citet{roy2020weakly} extended this work with subframe lexicons to refine the broad categories of the MFC. Some have considered the way Moral Foundations \cite{haidt2007morality} can serve as issue frames \cite{kwak2020frameaxis,mokhberian2020moral,priniski2021mapping}, and others have built issue-specific typologies \cite{mendelsohn2021modeling}. While issue framing has been well-studied \cite{ajjour2019modeling,baumer2015testing}, entity framing remains \emph{under-examined} in NLP with a few exceptions. One line of work used an unsupervised approach to identify \textsl{personas} or clusters of co-occurring verbs and adjective modifiers \cite{card2016analyzing,bamman2013learning,iyyer2016feuding,joseph2017girls}. Another line of work combined psychology lexicons with distributional methods to measure implicit differences in power, sentiment, and agency attributed to male and female entities in news and film \cite{sap2017connotation,field2019entity,field2019contextual}. 

Social scientists have experimentally manipulated framing devices related to police violence, including \textsl{law and order}, \textsl{police brutality}, or \textsl{racial stereotypes}, revealing dramatic effects on participants' perceptions of police shootings \cite{fridell2017explaining,dukes2017black,porter2018public}. Criminologists have trained coders to manually annotate on the order of 100 news articles for framing devices relevant to police use of force \cite{hirschfield2010legitimating,ash2019framing}. While these studies provide great theoretical insight, their manual coding schemes and small corpora are not suited for large scale real-time analysis of news reports nationwide. While many recent works in NLP have started to detect police shootings \cite{nguyen2018killed,keith2017identifying} and other gun violence \cite{pavlick2016gun}, we are the first to model entity-centric framing around police violence.

\section{\data{}}

\begin{table*}
    \centering
    \resizebox{\textwidth}{!}{
    \begin{tabular}{rrrrrrrrr}
\toprule
\textbf{Leaning} & \textbf{Articles} (\#) & \textbf{Events} (\#) & \textbf{Sources} (\#) & \textbf{Armed} (\%) & \textbf{Attack} (\%) & \textbf{Fleeing} (\%) & \textbf{Mental Illness} (\%) & \textbf{Video} (\%) \\ \midrule

Left & 1,090 & 6730 & 90 & 58.0 & 35.1 & 24.6 & 20.6 & 13.6 \\
Left Center & 9,761 & 3,794 & 233 & 66.1 & 45.9 & 23.8 & 18.0 & 11.1 \\
Least Biased & 5,214 & 3,428 & 164 & 73.3 & 53.7 & 26.2 & 19.3 & 8.8 \\
Right Center & 3,993 & 2,631 & 105 & 71.1 & 50.9 & 24.9 & 18.0 & 9.9 \\
Right & 1,009 & 782 & 40 & 64.8 & 48.7 & 23.3 & 19.3 & 15.5 \\
None & 59,739 & 7,300 & 12,931 & 71.0 & 52.3 & 24.9 & 18.4 & 8.9 \\ \hline
Total & 80,806 & 7,647 & 13,563 & 70.3 & 51.3 & 24.8 & 18.4 & 9.4 \\ \bottomrule
    \end{tabular}
    }
    \caption{\small \data{} statistics. The number of articles and the breakdown of events by whether the victim was \textit{Armed, Attacking, Fleeing}, had \textit{Mental Illness} or was filmed on \textit{Video} according to Mapping Police Violence data. \textit{Leaning} is decided via Media Bias Fact Check in  Section~\ref{susbsec:assigning_media_slant_labels}.}
    \label{tab:dataset_statistics}
\end{table*}

To study media framing of police violence, we introduce \data{} (\adata) which contains over 82,000 news reports of police shooting events. We now describe the corpus construction as it is shown in Figure \ref{fig:pipeline}.

\subsection{Identifying Shooting Events}

We first use Mapping Police Violence \cite{mpv2021} to identify shooting events. It is a representative, reliable, and detailed record, as it cross-references the three most complete databases available: Fatal Encounters \cite{fatalEncounters}, the U.S. Police Shootings Database \cite{policeShootingsDatabase}, and Killed by Police, all of which have been validated by the Bureau of Justice Statistics \cite{banks2016arrest}. Prior works in sociology and criminology \cite{gray2020race} use it as an alternative to official police reports because local police departments significantly underreport shootings \cite{williams2019limitations}.
At the time of our retrieval, the Mapping Police Violence dataset identified 8,169 named victims of police shootings between January 1, 2013 and September 4, 2020 and provided the victim's age, gender, and race, whether they fled or attacked the officer, and whether the victim had a known mental illness or was armed (and with what weapon), as well as  the location and date of the shooting, the agency responsible, and whether the incident was recorded on video.

\subsection{Collecting News Reports}

For each named police shooting or violent encounter in Mapping Police Violence, we query the Google search API for up to 30 news articles relevant to that event. We found this sample size is large enough to represent both sides without introducing too much noise. Our query string includes officer keywords, the victim's name, and a time window restricted to within one month of the event (see Appendix~\ref{sec:news_data_collection} for details and design choices). Next, we extracted article publication dates using the \texttt{Webhose} extractor \cite{geva_2018}, and as a preprocessing step, we used the \texttt{Dragnet} library \cite{peters2013content} to automatically filter and remove ads, navigation items, or other irrelevant content from the raw HTML. In the end, the \data{} contained 82,100 articles across 7,679 events. The per-ideology statistics of reported events are given in Table~\ref{tab:dataset_statistics}. The racial and ethnic distribution is: White (43.0\%), Black (29.7\%), Hispanic (15.3\%), Asian (1.5\%), Native American (1.3\%), and Pacific Islander (0.5\%), while the other 8.7\% of articles report on a victim of unknown race/ethnicity.

\subsection{Assigning Media Slant Labels}
\label{susbsec:assigning_media_slant_labels}

We associated each news source with a political leaning by matching its URL domain name with the \citet{mbfc2020} record. With more than 1,500 records, the MBFC contains the largest available collection of crowdsourced media slant labels, and it has been used as ground truth in other recent work on news bias \cite{dinkov2019predicting,baly2018predicting,baly2019multi,nadeem2019fakta,stefanov2020predicting}. The MBFC labels are \textsl{extreme left}, \textsl{left}, \textsl{left-center}, \textsl{least biased}, \textsl{right-center}, \textsl{right}, and \textsl{extreme right}. For our political framing analysis (Section~\ref{sec:partisan_framing_analysis}), we consider a source liberal if its MBFC slant label is \textsl{left} or \textsl{extreme left}, and we consider the source conservative if its label is \textsl{right} or \textsl{extreme right}. We manually filter this polarized subset to ensure that all articles are on-topic. This led to 1,090 liberal articles and 1,002 conservative articles. 

\section{Media Frames Extraction}
\label{sec:extracting}

We are interested in both the issue and entity frames that structure public narratives on police violence. We will now present our computational framework for extracting both from news text. Throughout this section, we cite numerous prior works from criminology and sociology to motivate our taxonomy, but we are the first to measure these frames computationally. As a preview of the system, Figure~\ref{fig:frame_example} shows the key frames extracted from two articles on the murder of 15-year-old Jordan Edwards. Notably, only the left-leaning article mentions the victim's \textsl{race}. Most importantly, our system distinguishes the victim's attributes from descriptions of the officer. Here, it is the \textsl{officer} who is described as a ``killer,'' and not the victim.

\subsection{Entity-Centric Frames}
\label{sec:entity_frames}

Our entity-centric analysis and lexicons are a key contribution in this work. We distinguish the victim's attributes like \textsl{race} and \textsl{armed} status from that of the officer or some other entity, and so we move beyond generic and global \textsl{issue} frames to understand how the target population is portrayed. These methods require a partitioning of entity tokens into \textsc{victim} and \textsc{officer} sets. To do so, we first append to each set any tokens matching a \textsl{victim} or \textsl{officer} regex. The officer regex is general, but the victim regex matches the known name, race, and gender of the victim in \adata, like \textit{Ronette Morales, Hispanic, woman} (See Appendix~\ref{sec:frame_extraction_methods}). Second, we use the huggingface \texttt{neuralcoref}
for coreference resolution based on \citet{clark2016deep}, and append all tokens from spans that corefer to the \textsc{victim} or \textsc{officer} set respectively.

\textbf{Age, Gender, and Race.} Following \citet{ash2019framing}, we consider the age, gender and race of the victim, which are central to an intersectional understanding of unjust police conduct \cite{dottolo2008don}. We extract age and gender frames by string matching on the gender modifier or the numeric age. We extract race frames by searching the attributive or predicative adjectives and predicate nouns of \textsc{victim} tokens and matching these with the victim's known race.

\textbf{Armed or Unarmed.} Knowing whether the victim was armed or unarmed is a crucial variable for measuring structural racism in policing \cite{mesic2018relationship}. We identify mentions of an unarmed victim with the regex \texttt{unarm(?:ed|ing|s)?}, and mentions of an armed victim with \texttt{arm(ed|ing|s)?}, excluding tokens with noun part-of-speech.

\textbf{Attacking or Fleeing.} Since \textsl{Tennessee v. Garner} (1985), the lower courts have ruled that police use of deadly force is justified against felons in flight only when the felon is dangerous \cite{harmon2008police}. Since \textsl{Plumhoff v. Rickard} (2014), deadly force is justified by the risk of the fleeing suspect. Thus whether the victim fled or attacked the officer can inform the officer's judgment on the appropriateness of deadly force. We propose an entity-specific string-matching method to extract attack frames, where a \textsc{victim} token must be the head of a verb like \textit{injure}, and we and use expressions like \texttt{\textbackslash bflee(:?ing)} to extract fleeing mentions.

\textbf{Criminality.}
Whether the article frames the victim as someone who has engaged in criminal activity may serve to justify police conduct \cite{hirschfield2010legitimating}. To capture this frame, we used Empath \cite{fast2016empath} to build a novel lexicon of unambiguously criminal behaviors (e.g. \textsl{cocaine}, \textsl{robbed}), and searched for these terms. 

\textbf{Mental Illness.} Police are often the first responders in mental health emergencies \cite{patch1999police}, but there is growing concern that the police are not sufficiently trained to de-escalate crisis situations involving persons with mental illness \cite{kerr2010police}. Mentioning a victim's mental illness may also highlight evidence of this structural shortcoming. We again used Empath to build a custom lexicon for known mental illnesses and their correlates (e.g. \textsl{alcoholic}, \textsl{bipolar}, \textsl{schizophrenia}). As for \textit{Criminality}, this is not an exhaustive list; we balance precision and recall by ensuring that terms are unambiguous in the context of police violence. Still, we may not capture other signs of mental illness, like descriptions of erratic behaviors.

\subsection{Issue Frames}
\label{sec:issue_frames}
\hspace{\parindent}\textbf{Legal Language.} Similar to \citet{ash2019framing}, we investigate frames which emphasize legal outcomes for police conduct. To capture this frame, we used a public lexicon of legal terms from the Administrative Office of the U.S. Courts \citeyear{uscourts}.

\textbf{Official and Unofficial Sources.} Many news accounts favor official reports which frame police violence as the state-authorized response to dangerous criminal activity \cite{hirschfield2010legitimating,lawrence2000politics}. Others may include unofficial sources like interviews with first-hand witnesses. We identify official and unofficial sources with the following Hearst-like patterns: $<$source$>$ $<$verb$>$ $<$clause$>$ or \textit{according to} $<$source$>$, $<$clause$>$. Our unique entity-centric approach lets us exclude the victim's quotes and focus on witness testimony.

\textbf{Systemic.} While the news has historically favored episodic \cite{iyengar1990framing} fragmented \cite{bennett2016news}, or decontextualized narratives \cite{lawrence2000politics}, there has been an increase in systemic framing since the 1999 shooting of Amadou Diallo \cite{hirschfield2010legitimating}. Such articles identify police shootings as the product of structural or institutional racism. To extract this frame, we look for sentences that (1) mention \textsl{other} police shooting incidents or (2) use keywords related to the national or global scope of the problem. 

\textbf{Video.} Video evidence was a catalyst for the Rodney King protests \cite{lawrence2000politics}. Psychology studies have found that subjects who witnessed a police shooting on video were significantly more likely to consider the shooting unjustified compared with those who observed through news text or audio \cite{mccamman2017police}. We identify reports of body or dash camera footage using the simple regex \texttt{(body(?: )?cam|dash(?: )?cam)}

\subsection{Moral Foundations}

Moral Foundations Theory \cite{haidt2007morality} (MFT) is a framework for understanding universal values that underlie human judgments of right and wrong. These values form five dichotomous pairs: care/harm, fairness/cheating, loyalty/betrayal, authority/subversion, and purity/degradation. While MFT is rooted in psychology, it has since been applied in political science to differentiate liberal and conservative thought \citet{graham2009liberals}. We quantify the moral foundations that media invoke to frame the virtues or vices of the officer and the victim in a given report using the extended MFT dictionary of \citet{rezapour2019enhancing}.

\subsection{Linguistic Style}
To supplement our understanding of overtly topical \textsl{entity} and \textsl{issue} frames, we investigate two relevant linguistic structures: passive verbs and modals. 

\textbf{Passive Constructions.} Prior works identify framing effects that arise from passive phrases in narratives of police violence \cite{hirschfield2010legitimating,ash2019framing}. In this work, we distinguish \textsl{agentive} passives (e.g. ``He was killed \textsl{by} police.'') from \textsl{agentless} passives (e.g. ``He was killed.''). While both deprive actors of agency \cite{richardson2006analysing}, only the latter obscures the actor entirely, effectively removing any blame from them \cite{greene2009more}. We specifically contrast liberal and conservative use of \textsc{victim}-headed agentless passives (passive verbs whose patient belongs to the \textsc{victim} set).

\textbf{Modal Verbs.} Modals are used \textsl{deontically} to express necessity and possibility, and in this way, they are often used to make moral arguments, suggest solutions, or assign blame \cite{portner2009modality}. Following \citet{demszky2019analyzing}, we count the document-level frequency of tokens belonging to four modal categories: MUST, SHOULD (\textit{should / shouldn't / should've}), NEED and HAVE TO. 

\section{Validating Frame Extraction Methods}
\label{sec:validation}

One coder labeled 50 randomly-sampled news articles with indices to mark the order of frames present. Against this ground truth, our binary frame extraction system achieves high precision and recall scores above 70\%, with only \textsl{race} and \textsl{unofficial sources} at 66\% and 65\% precision. Accuracy is no less than 70\% for any frame (see Table~\ref{tab:extraction_performance} in Appendix~\ref{appdx:validation}). One advantage of our system is that it is not a black box -- it gives us the precise location of each frame in the document. When we sort the indices of the predicted frame locations, we find that our system achieves a 0.752 Spearman correlation with the ground truth frame ordering. Finally, the annotator gave us a rank order of the officer and victim moral foundations most exemplified in the document. When we sort, for each document, the foundations by score, our system achieves 0.66 mAP for the officer and 0.40 mAP for the victim.

\section{Political Framing of Police Violence}
\label{sec:partisan_framing_analysis}

\subsection{Frame Inclusion Aligns with Slant} 
\label{subsec:cat_freq}

As shown in  Table~\ref{tab:partisan_framing} (\textit{Left}), we find that liberal sources frame the issue of police violence as more of a systemic issue, using \textsl{race}, \textsl{unarmed}, and \textsl{mental illness} entity frames, while conservative sources frame police conduct as justified with regard to an uncooperative victim. 

Specifically, conservative sources more often mention a victim is \textsl{armed} (.588 vs. .439, $+34\%$), %, Cohen's d=$.302$
\textsl{attacking} ($+46\%$), % $d=.347$
and \textsl{fleeing} ($+47\%$). % , $d=.244$). 
These strategies serve to justify police conduct since \textsl{Tennessee v. Garner} (1985) affirmed the use of deadly force on dangerous suspects in flight \cite{harmon2008police}, and this narrative is furthered by \textsl{official sources} ($+38\%$), %, $d=.496$
\textsl{legal language} ($+5\%$), %, $d=.144$
and the victim's \textsl{criminal record} ($+7\%$). %, $d=.087$
Liberal news instead emphasizes the victim's \textsl{race} ($+100\%$), %, $d=.471$
\textsl{mental illness} ($+25\%$), %, $d=.097$
and that the victim was \textsl{unarmed} 
($+25\%$). %, $d=.239$
Cumulatively, these details reinforce the prominent \textsl{systemic racism} narrative that appears $47\%$ more often in liberal media. %($d=.286$)
The victim's mental illness may signal police failure to handle mental health emergencies \cite{kerr2010police}, and the police killing of an unarmed Black victim provides evidence of institutional racism in law enforcement \cite{aymer2016can,tolliver2016police}. Together with gender and age, the victim's race informs an intersectional account of police discrimination \cite{dottolo2008don}. Surprisingly, we find that liberal sources mention age and gender significantly \textsl{less} often than do conservative sources. We find no significant differences in the mention of video evidence, possibly because this detail is broadly newsworthy.  

\textbf{Controlling for confounds amplifies ideological differences.} One potential confound is \textsl{agenda setting}, or ideological differences in the amount of coverage that is devoted to different events \cite{mccombs2002agenda}. In fact, conservative sources were significantly more likely to cover cases in which the victim was \textsl{armed} and \textsl{attacking} overall. However, our findings in this section are actually \textbf{magnified} when we level these differences and consider only news sources where the ground truth metadata reflects the framing category (see Appendix~\ref{appdx:supp_framing_analysis}).

\begin{figure*}
\centering
    \includegraphics[width=0.9\textwidth]{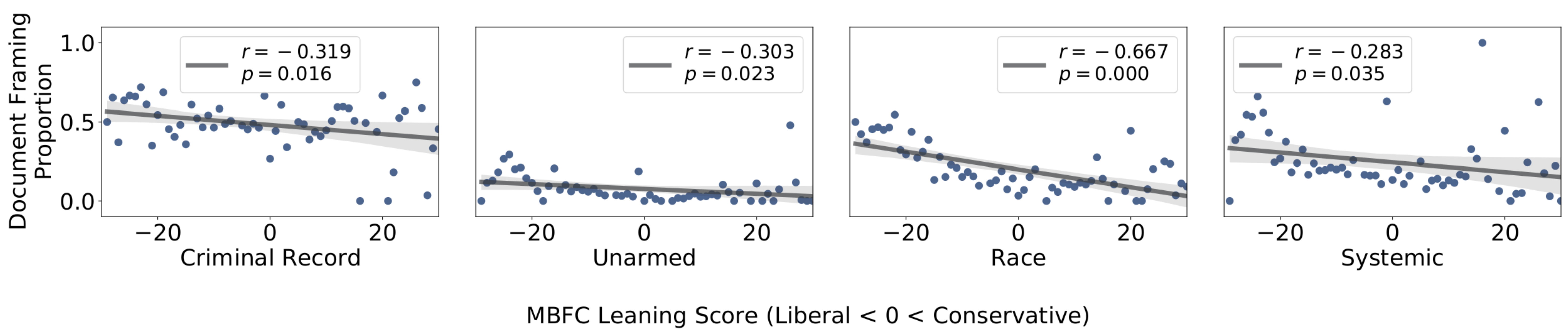}
    \caption{\small \textbf{Framing as a function of political leaning.} The MBFC political leaning score vs. the document frame proportion. 
    }\label{fig:reg}
\end{figure*}

\textbf{Framing decisions are a \textit{function} of slant.} Finally, we expect that news sources will differ not only diametrically at the political poles, but also linearly in the degree of their polarization. To examine this, we collected an integer score ranging from -35 (extreme left) to +35 (extreme right), which we scraped from the MBFC using an open source tool \cite{carey2021}. We aggregated articles by their political leaning scores and found the proportion of articles in each bin that express the frame. Linear regressions in Figure~\ref{fig:reg} reveal a statistically significant negative correlation between conservatism and the \textsl{criminal record} (r=-0.319), \textsl{unarmed} (r=-0.303), \textsl{race} (r=-0.667) and \textsl{systemic} frames (r=-0.283).

\begin{table}
    \centering
    \resizebox{\columnwidth}{!}{
    \begin{tabular}{lllccllc}
         \toprule
         & \multicolumn{3}{c}{\textbf{Inclusion}} & \phantom{a} & \multicolumn{3}{c}{\textbf{Ordering}} \\
         \cmidrule{2-4} \cmidrule{6-8} \\
         \textbf{Framing Device} & \textbf{Lib.} & \textbf{Cons.} & $p$ & & \textbf{Lib.} & \textbf{Cons.} & $p$ \\ \midrule 
        Age & 0.472 & \textbf{0.764} & $^{***}$ & & \textbf{0.480} & 0.467 & \\
        Armed & 0.439 & \textbf{0.588} & $^{***}$ & & 0.313 & \textbf{0.358} & $^*$\\
        Attack & 0.369 & \textbf{0.539}  & $^{***}$ & & 0.267 & \textbf{0.306} & \\
        Criminal record & 0.613 & \textbf{0.655} & $^{*}$ & & \textbf{0.294} & 0.278 & $^{**}$ \\
        Fleeing  & 0.228 & \textbf{0.336}  & $^{***}$ & & \textbf{0.246} & 0.217 &\\
        Gender & 0.610 & \textbf{0.620} & $^{***}$ & & 0.611 & \textbf{0.622} & \\
        Legal language & 0.875 & \textbf{0.919} & $^{***}$ & & \textbf{0.523} & 0.419 & $^{***}$ \\
        Mental illness & \textbf{0.181} & 0.145 & $^{*}$ & & 0.301 & 0.296 & \\
        Official sources & 0.586 & \textbf{0.808} & $^{***}$ & & \textbf{0.194} & 0.163 & $^{***}$ \\
        Race & \textbf{0.428} & 0.214 & $^{***}$ & & \textbf{0.296} & 0.233 & $^{***}$ \\
        Systemic & \textbf{0.428} & 0.291 & $^{***}$ & & \textbf{0.410} & 0.283 & $^{***}$ \\
        Unarmed & \textbf{0.195} & 0.110 & $^{***}$ & & 0.408 & \textbf{0.470} & \\
        Unofficial sources  & 0.708 & \textbf{0.780} & $^{**}$ & & \textbf{0.184} & 0.163 & $^{***}$\\
        Video  & 0.164 & \textbf{0.191} & & & 0.283 & \textbf{0.436} & $^{***}$\\
         \bottomrule 
    \end{tabular}
    }
    \caption{\small (\textit{Left}) \textbf{Frame inclusion aligns with political slant.} The proportion of liberal and conservative news articles that include the given framing device. (\textit{Right}) \textbf{Frame ordering aligns with media slant.} The average inverse document frame order in liberal and conservative news articles where the \textsl{frame is present}.
    Significance given by Mann-Whitney rank test:  * ($p<0.05$), ** ($p<0.01$), *** ($p<0.001$)}
    \label{tab:partisan_framing}
\end{table}

\subsection{Frame Ordering Aligns with Slant}
\label{subsec:frame_ordering}

The Inverted Pyramid, one of the most popular styles of journalism, dictates that the most important information in an article should come first \cite{po2003news,upadhyay2016making}. We hypothesize that the ordering of frames will reflect the author's judgment on which details are most important, so we should observe ideological differences in frame ordering. In Table~\ref{tab:partisan_framing} (\textit{Right}) we compare, for each frame, its average inverse document rank in liberal and conservative news articles in which the frame was \textsl{already present}. We find that conservative sources highlight that the victim is armed and attacking by placing these details earlier in the report when they are mentioned (avg. inverse rank .358 vs. .313 for \textsl{armed}, 
%Cohen's $d=.166$ 
and .306 vs. .267 for \textsl{attacking}%, $d=.164$
). By prioritizing these details early in the article, conservative sources further highlight the need for law and order \cite{drakulich2020race,fridkin2017race}.

Although in Section~\ref{subsec:cat_freq} we found conservative sources favored police reports, we now observe a liberal bias favoring early quotations from these \textsl{official sources} (.194 vs. .163 avg. inverse rank
%, $d=.256$
). At the same time, liberal sources highlight \textsl{unofficial sources} like eyewitnesses who may identify police brutality as a ``pervasive and endemic problem'' \cite{lawrence2000politics}. Most notably, liberal sources prioritize \textsl{legal language} (0.523 vs. 0.419)
%, $d=.324$
and \textsl{systemic} framing (0.410 vs. 0.283).
%, $d=.393$
Liberal sources place these frames, on average, \textsl{second} in the total frame ordering (inverse rank $\approx 1/2$), which primes readers to interpret almost all other remaining frames through the lens of injustice and structural racism. This confirms prior work \cite{graham2009liberals,hirschfield2010legitimating}. 

\begin{figure*}
\centering
    \includegraphics[width=1.0\textwidth]{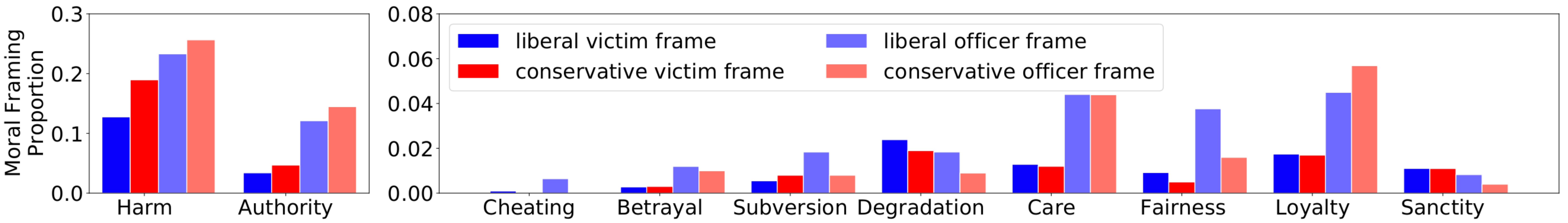}
    \caption{\small \textbf{Differences in moral foundation frames.} The average moral framing proportions in liberal and conservative articles.
    }
    \label{fig:mft_results}
\end{figure*}

\subsection{Moral Framing Differences}
\label{subsec:moral_framing_differences}

Prior work \cite{graham2013moral,haidt2007morality} suggests that liberals emphasize the care/harm and fairness/cheating dimensions, especially as vice in the officer \cite{lawrence2000politics}, while conservatives might defend the foundations more equally, especially as virtue in the officer or vice in the victim \cite{drakulich2020race}. We test this by computing, for each moral foundation and for each entity (victim, officer), the proportion of liberal and conservative articles in which either a modifier or agentive verb from the \citet{rezapour2019enhancing} moral foundation dictionary is used to describe that entity. Figure~\ref{fig:mft_results} shows that conservative sources unsurprisingly place more emphasis on the victim's \textsl{harm}ful behaviors ($+48\%$
%, $d=.171$
). 
Only liberal sources mention the officer's unfairness or \textsl{cheating}. 
%($d=.111$). 
Liberal articles also include more mentions of officer \textsl{subversion} ($+130\%$
%, $d=.09$
) and, surprisingly, \textsl{fairness} ($+135\%$
%, $d=.133$
). These results are all significant with $p<0.05$; no other ideological differences are significant.

\subsection{The Politics of Linguistic Style}
Liberal politicians largely support Black Lives Matter and its calls for police reform \cite{hill2018crime}, while conservative politicians have historically opposed the Black Lives Matter movement or any anti-police sentiment \cite{drakulich2020race}. We hypothesize that
there will be significant differences in the use of agentless passive constructions and modal verbs of necessity \cite{greene2009more,portner2009modality} between conservative and liberal sources. When we compare the average document-level frequency for each framing device, normalized by the length of the document in words, we find these hypotheses supported in Table~\ref{tab:linguistic_style}. Liberal sources use modal verbs of necessity like SHOULD and HAVE TO more frequently. Conservative sources use agentless passive constructions 61\% more than liberal sources (2.55$\times 10^{-3}$ vs. 1.58$\times 10^{-3}$), and \textsl{violent} passives\footnote{To indicate violence, we check that the lemma is in \{`attack', `confront', `fire', `harm', `injure', `kill', `lunge', `murder', `shoot', `stab', `strike'\}} 31\% more. However, we also find that conservative sources discuss the victim more overall ($+34\%$%, $d=.27$
). To remove this confound, we re-normalize the \textsl{Passive} and \textsl{Violent Passive} counts by the number of victim tokens instead, and the results still hold: $+39\%$ and $+22\%$ respectively. 

\section{The Broader Scope of Media Framing}

\begin{figure*}
\centering
\includegraphics[width=0.9\linewidth]{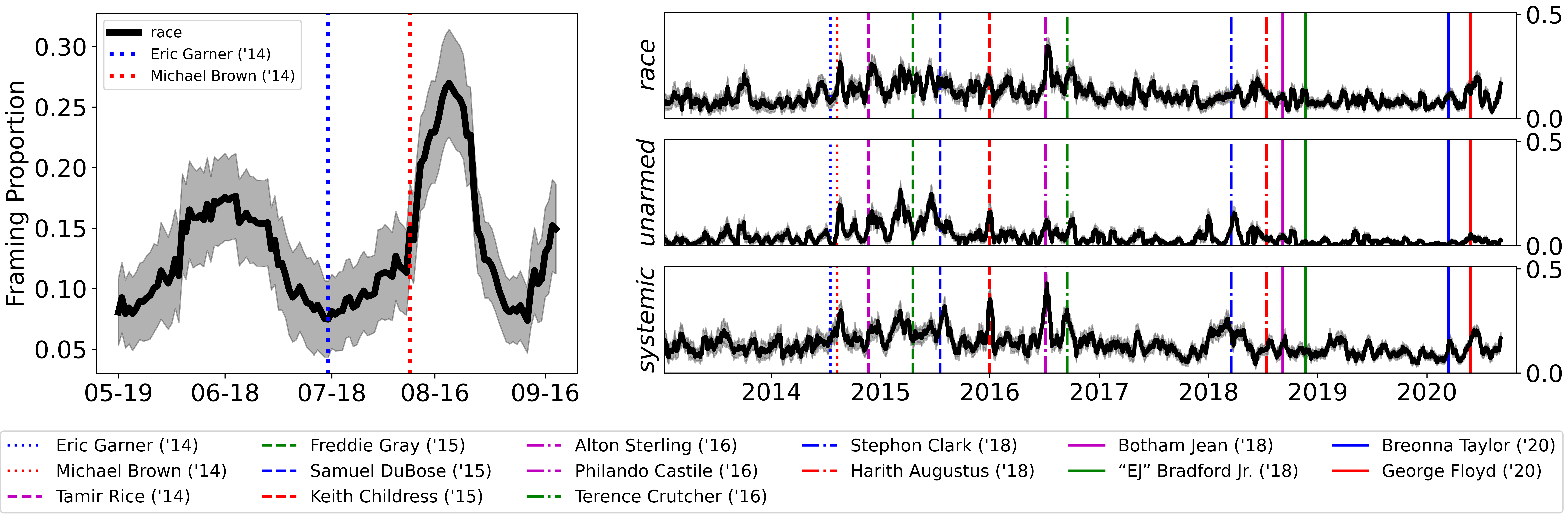}
    \caption{\small \textbf{Peaks near high-profile shootings.} Per-day proportion of articles with \textsl{race, unarmed,} and \textsl{systemic} frames included across time, excluding articles for the high-profile police shootings listed. This reveals local spikes near 15 high-profile incidents. For example, on the \textit{Left} we see \textsl{race} framing spikes after the death of Michael Brown.
    }
    \label{fig:temporal_analysis}
\end{figure*}

Prior works have algorithmically tracked collective attention in the news cycle \cite{leskovec2009meme}, and measured the correlation between media attention and offline political action \cite{de2016social,holt2013age,mooijman2018moralization}. This section examines the broader scope of media framing via two studies.

\subsection{Peaks Near High-Profile Killings}
\label{sec:temporal_peaks}

Do news framing strategies co-evolve across time? We expect to see coordinated peaks in the prevalence of \textsl{race}, \textsl{unarmed}, and \textsl{systemic} frames across U.S. news media, especially near high-profile killings of unarmed Black American citizens. To investigate this hypothesis, we took, for each salient frame, the proportion of articles that mention that frame out of the 82,000 news articles in our dataset, \textsl{excluding} any articles that report one of the 15 high-profile police killings listed in Figure~\ref{fig:temporal_analysis}. Then we found the Pearson correlation between each of the time series in a pairwise manner: 0.49 systemic/unarmed, 0.56 race/unarmed, and 0.70 race/systemic; all statistically significant with $p<2.0\times 10^{-167}$.

\begin{table}[t]
    \centering
    \small
    \resizebox{\columnwidth}{!}{
    \begin{tabular}{L{2.2cm}rrrr}
         \toprule
         \textbf{Framing Device} & \textbf{Lib.} & \textbf{Cons.} & \textbf{Cohen's \textsl{d}} \\ \midrule 
    MUST ** & \textbf{2.09e-4} & 1.03e-5 & 0.132 \\ \midrule
    SHOULD *** & \textbf{4.44e-4} & 2.04e-4 & 0.262 \\ \midrule
    NEED *** & \textbf{2.64e-4} & 1.15e-4 & 0.190 \\ \midrule
    HAVE TO *** & \textbf{4.09e-4} & 1.75e-4 & 0.196 \\ \midrule
    Passive *** & 1.58e-3 & \textbf{2.55e-3} & 0.367 \\ \midrule
    Passive \newline Violence *** & 6.21e-4 & \textbf{8.13e-4} & 0.121\\
         \bottomrule 
    \end{tabular}
    }
    \caption{\small \textbf{Politics and linguistic styles.} The average document frequency of linguistic structures, normalized by the length of the document. $^{***}$ $p<0.001$
    }
    \label{tab:linguistic_style}
\end{table}

The time series in Figure~\ref{fig:temporal_analysis}, smoothed over a 15-day rolling window, reveal local spikes near each of the high-profile killings, with the largest surge in racial and systemic framing near the shootings of \textsc{Alton Sterling} and \textsc{Philando Castile}. Two of the earliest surges appear near the killing of \textsc{Eric Garner} and \textsc{Michael Brown}, which largely ignited the Black Lives Matter movement \cite{carney2016all}. Recent spikes also appear near the killing of \textsc{Breonna Taylor} and \textsc{George Floyd}, which sparked record-setting protests in 2020  \cite{buchanan2020black}. We quantify this with an intervention test \cite{toda1995statistical} on each time series $X = (X_1, X_2, ..., X_t, ...)$ by fitting an AR(1) model defined by
\begin{equation*}
    X_t = \beta_0 X_{t-1} + \beta_1 P(t)  + c + \epsilon_t
\end{equation*}
with parameters $\beta$, constant $c$, error $\epsilon_t$, and a pulse function $P(t)$ to indicate the intervention
\begin{equation*}
    P(t) = \begin{cases}
        1, & \text{there was a high-profile shooting at $t$} \\
        0, & \text{else} 
    \end{cases}
\end{equation*}
The AR(1) is an auto-regressive model where only the previous term $X_{t-1}$ influences the prediction for $X_t$, and the intervention $P(t)$ allows us to test the null hypothesis that a high-profile killing does not impact framing proportions ($\beta_1 = 0$). We find the coefficient on the intervention $\beta_1$ is positive for each frame, and significant only in the \textsl{unarmed} regression ($\beta_1=2.03$, $p<0.01$). Given this and the high correlation between the three framing categories, we conclude that high-profile killings influence media decisions to frame other killings.

\subsection{Political Action Precedes Media Framing}
\label{sec:predicting_political_action}

We predict that protest volume will positively correlate with media attention on the race and unarmed status of recent victims and the underlying systemic injustice of police killings. Using the CountLove \cite{leung_perkins_2021} protest volume estimates from January 15, 2017 through December 1, 2020, we aligned the per-day national volume with the \textsl{race}, \textsl{unarmed}, and \textsl{systemic} time series (Figure~\ref{fig:temporal_analysis}) and found low but positive Pearson correlations of 0.098, 0.073, and 0.088 respectively, each statistically significant. These correlations were directed, with protest volume \textsl{Granger-causing} an increase in these framing strategies. For two aligned time series $X$ and $Y$, we say $X$ \textsl{Granger-causes} $Y$ if past values $X_{t-\ell} \in X$ lead to better predictions of the current $Y_t \in Y$ than do the past values $Y_{t-\ell} \in Y$ alone \cite{granger1980testing}. Here, $\ell$ is called the lag. We considered a lag of 1 and a lag of 2 days. The rightmost column of Figure~\ref{fig:temporal_analysis} shows that, with $\ell=2$, protest volume Granger-causes  \textsl{race} and \textsl{unarmed} framing with statistical significance $(p<0.05)$ by the SSR F-test. The reverse direction is \textsl{not} statistically significant. This reveals that offline protest behaviors \textsl{precede} these important media framing decisions, not the other way around. This echoes similar findings on media shifts after the Ferguson protests \cite{arora2019framing} and social media engagement after protests like Arab Spring \cite{wolfsfeld2013social}.

\begin{table}
    \centering
    \small
    \resizebox{\columnwidth}{!}{
    \begin{tabular}{L{1.2cm}L{1.5cm}|R{1.5cm}R{1.5cm}}
         \toprule
         \textbf{Frame} & Pearson $r$ & Granger \newline 1-lag $p$ & Granger \newline 2-lag $p$ \\ \midrule
         Race & 0.098 & 0.0185 & 0.0381 \\
         Unarmed & 0.073 & 0.1646 & 0.0024 \\
         Systemic & 0.088 & 0.0801 & 0.0600\\
         \bottomrule
    \end{tabular}
    }
    \caption{\small{\textbf{Media Framing and Political Action.} Correlation and $p$ values for political protests \textit{Granger-causing} media attention towards the race, unarmed, and systemic frames.}}
    \label{tab:granger_causality}
\end{table}

\section{Discussion and Conclusion}
In this work, we present new tools for measuring entity-centric media framing, introduce the  \data{}, and use them to understand media coverage on police violence in the United States. Our work  uncovers 15 domain-relevant framing devices and reveals significant differences in the way liberal and conservative news sources frame both the \textsl{issue} of police violence and the \textsl{entities} involved. We also show that framing strategies co-evolve, and that protest activity precedes or anticipates crucial media framing decisions. 

We should carefully consider the limitations of this work and the potential for bias. Since we matched age, gender and race directly with the MPV, we expect minimal bias, but acknowledge that our exact string-matching methods will miss context clues (e.g. drinking age), imprecise referents (e.g. ``teenager''), and circumlocution. We also rely on lexicons derived from expert sources (e.g. U.S. Courts \citeyear{uscourts}) or from data (Empath), both of which are inherently incomplete. Even the most straightforward keywords (e.g. armed, fleeing) are prone to error. However, biases could also appear in discriminative text classifiers. The advantage of our approach is that it is interpretable and extractive, allowing us to identify matched spans of text and quantify differences in frame ordering. Furthermore, it is grounded heavy in the social science literature. Similar methods could be applied to other major issues such as climate change \cite{luo2020desmog} and immigration  \cite{mendelsohn2021modeling}, where entities include politicians, protesters, and minorities, and where race, mental illness, and unarmed status may all be salient framing devices (e.g. describing the perpetrator or victim of anti-Asian abuse or violence; \citealt{gover2020anti, chiang2020anti, ziems2020racism, vidgen2020detecting}). 

\section*{Acknowledgments}
The authors would like to thank the members of SALT lab and the anonymous reviewers for their thoughtful feedback.

\section*{Ethical Considerations}
To respect copyright law and the intellectual property, we withhold full news text from the public data repository. Outside of the victim metadata, \data{} does not contain private or sensitive information. We do not anticipate any significant risks of deployment. However, we caution that our extraction methods are fallible (see Section~\ref{sec:validation}).

\bibliographystyle{acl_natbib}

\include{appendix}

\end{document}

%% file: appendix.tex
\appendix

\section{News Data Collection Methods}
\label{sec:news_data_collection}

Each news article search was made as a request to the Google search API using the following form

\begin{lstlisting}
https://www.google.com/search?q=q&num=30&hl=en
\end{lstlisting}

%.format(escaped_search_term, number_results+1, language_code)}

The query string \texttt{q} was structured in the following way. We included high-precision \textit{officer} and \textit{shooting} keywords, as well as the victim's full name string (which may contain a middle name or initial), with \texttt{\textbf{first\_name}, \textbf{last\_name}}), the first and last space separated word in the full name field respectively. We restricted the search to recent articles within one month following \texttt{\textbf{date}}, or the the day of the shooting event. We also included articles published on \texttt{\textbf{day}-days(1)} to account for possible time zone misalignment or imprecision.

\texttt{q = (\textbf{full\_name} OR \textbf{first\_name} OR \textbf{last\_name}) AND (shooting OR shot OR killed OR died OR fight OR gun) AND (police OR officer OR officers OR law OR enforcement OR cop OR cops OR sheriff OR patrol) after: \textbf{date}\texttt{-days(1)} before:\textbf{date} \texttt{+days(30)}}

The query returns up to 30 articles, which is equivalent to the first page of Google search results in a browser. We found this sample size of 30 to be large enough to contain a sufficient degree of diversity representing both liberal and conservative articles. A larger sample size could introduce additional noise or false positives in this data collection process.

% \subsection{Mapping Police Violence Dataset}
% % build a table
% \begin{table}[]
%     \centering
%     \small
%     \resizebox{\columnwidth}{!}{
%     \begin{tabular}{lR{40mm}R{50mm}}
%          \toprule
%          \textbf{Field} & \textbf{Description} & \textbf{Examples} \\ \midrule
%          Address & \\
%          Age & hammer, hatchet, knife, gun, etc. \\
%          Armed & armed, unarmed, unclear, vehicle
%  \\
%          \bottomrule
%     \end{tabular}
%     }
%     \caption{x}
%     \label{tab:mpv}
% \end{table}

\paragraph{Potential Confounds}
\label{appdx:confounds}
We are aware of some potential confounds in our data collection that could impact results. Firstly, some sources may not mention victim's name, and these articles will not be represented in our dataset. Articles that omit the victim's name may be particularly pro-police. Second, liberal and conservative sources could differ in their rate of publishing editorials, opinion pieces, or other content that is not strictly news-related. To investigate, one annotator labeled 100 randomly selected articles, 50 from the left and 50 from the right, indicating whether the article was news, opinion, or other. With simple binomial test, however, we just fail to reject the null hypothesis that the proportion of opinion pieces is statistically different between liberal and conservative sources (0.18 lib. vs. 0.06 cons., p=0.0.0648).

\section{Frame Extraction}
\label{sec:frame_extraction_methods}

Here we detail our frame extraction methods which come in two varieties. The first variety includes document-level regular expressions, and the second variety involves conditional string matching algorithms that rely on a partitioning of the all entity-related tokens into \textsc{victim} and \textsc{officer} sets. These extractive methods were ``debugged'' in minor ways after investigating their behavior on a development set, correcting for unexpected false positives and false negatives, but we did not iteratively refine regexes or extraction procedures to maximize precision and recall. Because our methods all extract spans of text, we were also able to verify that these rules were capturing different underlying segments of text. When we compute, for each pair of frames, the proportion of articles in which the difference between respective frame indices was within 25 tokens, we find the highest overlap between \textsl{legal language} and \textsl{criminal record} (25.5\%). However, only 10/91 pairs have >10\% overlap.

\subsection{Victim and Officer Partitioning}
First, we append to each set any tokens matching a \textit{victim} or \textit{officer} regex respectively. The victim regex matches the known name, race, and gender of the victim in the \adata dataset. For example, for the hispanic female victim named Ronette Morales, we would match tokens in the set 
\begin{lstlisting}
{`daughter', `female', `girl',
 `hispanic', `immigrant', 
 `latina', `latino', `mexican', 
 `mexican-american', `morales', 
 `mother', `ronette', `sister', 
 `woman'}
\end{lstlisting}

The officer regex, on the other hand, is given by
\begin{lstlisting}
    police|officer|\blaw\b|\benforcement\b|\bcop(?:s)?\b|sheriff|\bpatrol(?:s)?\b|\bforce(?:s)?\b|\btrooper(?:s)?\b|\bmarshal(?:s)?\b|\bcaptain(?:s)?\b|\blieutenant(?:s)?\b|\bsergeant(?:s)?\b|\bPD\b|\bgestapo\b|\bdeput(?:y|ies)\b|\bmount(?:s)?\b|\btraffic\b|\bconstabular(?:y|ies)\b|\bauthorit(?:y|ies)\b|\bpower(?:s)?\b|\buniform(?:s)?\b|\bunit(?:s)?\b|\bdepartment(?:s)?\b|agenc(?:y|ies)\b|\bbadge(?:s)?\b|\bchazzer(?:s)?\b|\bcobbler(?:s)?\b|\bfuzz\b|\bpig\b|\bk-9\b|\bnarc\b|\bSWAT\b|\bFBI\b|\bcoppa\b|\bfive-o\b|\b5-0\b|\b12\b|\btwelve\b
\end{lstlisting}

Second, we run the huggingface \texttt{neuralcoref} 4.0 pipeline for coreference resolution, and append all tokens from spans with coreference to the \textsc{victim} or \textsc{officer} set respectively. As an additional plausibility check, we ensure that at least one token in the span is recognized as being \textit{human}. By human, we mean either a proper noun, pronoun, a token with \texttt{spaCy} entity type PERSON, or a token belonging to the set of ``People-Related'' nouns extracted in \citet{lucy2020content} using WordNet hyponym relations.

\subsection{Document-Level Regular Expressions}
For the following categories, we used regular expression methods, returning the index of the first regex match, which we later sort for our final frame ranking. For categories with a dedicated lexicon, we used an exact string matching regex over these words `\texttt{\textbackslash bword1\textbackslash b|\textbackslash bword2\textbackslash b|...}' to match \texttt{word1}, \texttt{word2}, and all words in that lexicon. If no match was found, that framing category was said to be absent, and the frame rank was set to \texttt{inf}.

\subsubsection{Legal language} We compiled a lexicon of legal terms from the Administrative Office of the U.S. Courts \citeyear{uscourts},\footnote{https://www.uscourts.gov/glossary} supplemented with the Law \& Order terms listed in an online word list source.\footnote{http://www.eflnet.com/vocab/wordlists/law\_and\_order} We then hand-filtered any polysemous or otherwise ambiguous words like \textit{answer, assume,} and \textit{bench}, which could lead to false positives in a general setting. Finally, we employed an exact string matching regex over the words in the lexicon. 

\subsubsection{Mental illness.} To create a lexicon of terms related to mental illness, we used the Empath tool \cite{fast2016empath} to generate the words most similar to the token \texttt{mental\_illness} in an embedding space derived from contemporary New York Times data. We hand-filtered this set to remove generic illnesses and any words not related to mental health. We then employed an exact string matching regex over the words in the lexicon. 
% We copy the lexicon here.
% \begin{lstlisting}
% ['addicted', 'addiction', 'alcoholic', 'alcoholism', 'bipolar', 'dementia', 'depression', 'disability', 'disabled', 'emotional problems', 'mental disorder', 'mental health', 'mental illness', 'mental illnesses', 'mental problems', 'mental retardation', 'psychiatric problems', 'psychological problems', 'psychosis', 'retardation', 'schizophrenia', 'substance abuse']
% \end{lstlisting}

\subsubsection{Criminal record.} We again used Empath to create a lexicon of terms related to known crimes. We seeded the NYT similarity search with the terms \texttt{abuse}, \texttt{arson}, \texttt{crime}, \texttt{steal},  \texttt{trafficking}, and \texttt{warrant}. We then expanded this set using unambiguous crime names from the Wikipedia Category:Crimes page,\footnote{https://en.wikipedia.org/wiki/Category:Crimes} and finally hand-filtered so that the set included only crimes (e.g. \textit{theft}) or criminal substances (e.g. \textit{cocaine}). We then employed an exact string matching regex over the words in the lexicon. 
% We copy the lexicon here.
% \begin{lstlisting}
% ['abduct', 'abduction', 'abuse', 'apprehend', 'apprehended', 'arson', 'assault', 'break into', 'breaking into', 'broke into', 'burglary', 'burgle', 'cocaine', 'crack', 'crime', 'criminal', 'drug deal', 'drug dealing', 'drunk drive', 'drunk driving', 'fraud', 'guilty', 'homicide', 'kidnapping', 'manslaughter', 'misdemeanor', 'misdemeanors', 'molest', 'molestation', 'molested', 'offended', 'offender', 'perpetrate', 'perpetrators', 'pick pocket', 'prostitute', 'prostitution', 'warrant', 'rape', 'rapist', 'rob', 'robbed', 'robbery', 'sex crime', 'sex crimes', 'shoplift', 'shoplifted', 'shoplifting', 'speeding', 'stole', 'stolen', 'theft', 'thefts', 'thief', 'trafficking', 'trespassing', 'vandal', 'vandalism']
% \end{lstlisting}

\subsubsection{Fleeing.} To capture reports of a fleeing suspect, we use the following regular expression \texttt{(\textbackslash bflee(:?ing)?\textbackslash b|\textbackslash bfled\textbackslash b|\textbackslash bspe
(?:e)?d(?:ing)?(?:off|away|toward|
towards)|(took|take(:?n)?)off|
desert|(?:get|getting|got|run|
running|ran)away|pursu(?:it|ed))}. In this way, we identify fleeing both on foot (e.g. Minnesota 609.487, Subd. 6, \citeyear{mnlaw}) and via motor vehicle (e.g. California 2800.1 VC, \citeyear{calaw}). These are the forms of evasion that are explicitly enumerated by law. We include \texttt{pursu(?:it|ed)} to account for an evasion that is framed from the officer's perspective, which is a pursuit.

\subsubsection{Video.} We identify reports of body or dash camera footage using the simple regex \texttt{(body(?: )?cam|dash(?: )?cam)}. We do not use any other related lemmas like \texttt{video}, \texttt{film}, \texttt{record} because we found these to be highly associated with false positives, especially in web text where embedded videos are common. Similarly, we did not match on the word \texttt{camera} alone because of false-positives (e.g. ``family members declined on-camera interviews'').

\subsubsection{Age.}  According to the Associated Press Style Guide \cite{froke2019associated}, journalists should always report ages numerically. To avoid false positives, we do not match their spelled-out forms. We identify mention of age with an exact string match on the known numerical age of the victim, separated by \texttt{\textbackslash b} word boundaries.

\subsubsection{Gender.} Unlike \citet{sap2017connotation}), we are not interested in simply identifying the gender of the victim, but rather, whether there was specific mention of the victim's gender where a non-gendered alternative was available. For example, to avoid gendering a female victim, one could replace titles like \textit{mother} with \textit{parent}, \textit{daughter} with \textit{child}, \textit{sister} with \textit{sibling}, and \textit{female}, \textit{woman} or \textit{girl} with \textit{person} or simply with the name of the victim. Thus if the victim is female, we match \texttt{\textbackslash b(woman|girl|daughter|mother|
sister|female)\textbackslash b} and if the victim is male we match
\texttt{\textbackslash b(man|boy|son|father|
brother|male)\textbackslash b}. We do not match non-binary genders because we do not have ground truth labels for any non-binary targets.

\subsubsection{Unarmed}
We identify mentions of an unarmed victim with the regex \texttt{unarm(?:ed|ing|s)?}. Manual inspection of news articles reveals that this simple modifier is the standard adjective to describe unarmed victims, so it is sufficient in most cases. Unfortunately, it cannot capture other more subtle context clues (e.g. the victim was \textit{sleeping}, the victim's hands were in the air) or forms of circumlocuation. 

\subsubsection{Armed} We match individual tokens to the \texttt{\^{} arm(ed|ing|s)?} regex and only return the matching span for tokens that do not have a NOUN Part of Speech tag. This is necessary to disambiguate the verb \textit{arm} from the noun \textit{arm}. We can be confident that when an article mentions \textit{armed}, it is referring to the victim since an armed officer is not newsworthy. On the other hand, we do not match specific weapons because we cannot immediately infer that discussion about a weapon implies the \textit{victim} was armed (it could be an officer's weapon). We resolve this ambiguity when we extract \textsc{attack} frames, ensuring that the \textsc{victim} is the agent who is wielding a weapon object dependency.

\subsection{Matching Partitioned Tokens}
After partitioning the entity-related tokens into \textsc{victim} and \textsc{officer} sets, we extract the following frames for each document $\mathcal{D}$. In all of the following, we define the set \textsc{object} = \{\textit{dobj}, \textit{iobj}, \textit{obj}, \textit{obl}, \textit{advcl}, \textit{pobj}\} to indicate object dependencies.

\subsubsection{Race} 
We are determined to prune false positives from our \textit{race} frame detection. We only match \textit{race} where the \textit{race} term is given as an attributive or predicative modifier of the known victim.
To do so, we scan, for each token $t_k \in \textsc{victim}$, all children of the head of $t_k$ in the dependency parse. This set of children would include predicate adjectives of a copular head verb.  If the child matched with any member of the lexicon corresponding to the victim's race, we return the initial index of $t_k$. We also expect to capture adjective modifiers in this way because the \textsc{victim} tokens derive from entity spans that include modifiers.

\subsubsection{Attack.} 
Intuitively, we infer an article has mentioned an attack from the victim if we find the victim has acted in violence or has wielded an object that matches their known weapon or if the officer has been acted upon by a violent vehicular attack.
More specifically, for a given document, if we find an \textsc{victim} \textit{nsubj} token in that document having a verbal head in the \textsc{attack} set \{\textit{attack}, \textit{confront}, \textit{fire}, \textit{harm}, \textit{injure}, \textit{lunge}, \textit{shoot}, \textit{stab}, \textit{strike}\} or having a child with \textsc{object} dependency that matches the victim's known weapon type (e.g. \textit{gun}, \textit{knife}, etc.) then we return the token's index as an \textbf{attack} mention. To capture vehicular attacks, we also match tokens whose verbal head is in \{\textit{accelerate}, \textit{advance}, \textit{drive}\} and whose object is in the \textsc{officer} set. 
This process is detailed in Algorithm~\ref{alg:mentions_attack}, with a helper function in Algorithm~\ref{alg:verbs_with_obj}.

\begin{algorithm}
\SetAlgoLined
\KwIn{Dependency parsed document $\mathcal{D}$, and tokens $\mathcal{W}$ used to describe the victim's weapon (may be empty)}
\KwOut{Document string index $i$ of the token used to identify an attack from the victim}
\textsc{attack} $\leftarrow$ \{\textit{attack}, \textit{confront}, \textit{fire}, \textit{harm}, \textit{injure}, \textit{lunge}, \textit{shoot}, \textit{stab}, \textit{strike}\} \;
\textsc{advance} $\leftarrow$ \textit{accelerate}, \textit{advance}, \{\textit{drive}\} \;
\textsc{officer}, \textsc{victim} $\leftarrow$ \text{partition}$(\mathcal{D})$ \;
\For{$t_j \in \mathcal{D}$}{
    \If{dep$(t_j)=$ \textit{nsubj}}{
        \For{$(v, o) \in$ verbs\_with\_objs$(t_j, [\:])$}{ 
          \If{$(\text{lemma}(v) \in \textsc{attack})$ \textbf{and} $[(t_j \in \textsc{victim}) \: \textbf{or} \: (o \in \textsc{officer} \cup \mathcal{W}) ]$ }{
            \Return{index$(v, \mathcal{D})$}\;
             }
          \If{$v \in \textsc{advance}$ \: \textbf{and} \: $o \in \textsc{officer}$}{
            \Return{index$(v, \mathcal{D})$}\;
          }    
        }
    }
}
\Return{\texttt{inf}}\;
\caption{attack$(\mathcal{D}, \mathcal{W})$}
\label{alg:mentions_attack}
\end{algorithm}

\begin{algorithm}
\SetAlgoLined
\KwIn{Verb $v$ from dependency parsed document, recursively generated list $\mathcal{L}$ of (verb, object) tuples (initially empty) }
\KwOut{$\mathcal{L}$}
\For{$c \in \text{children}(v)$}{
    \If{$c \in \textsc{object}$}{
        append$((v,c), \mathcal{L})$\;
    }
    \ElseIf{dep$(c)$ = prep}{
        append$((v, \text{get\_pobj}(c)), \mathcal{L})$\;
    }
    \ElseIf{dep$(c) \in$ \{ \textit{conj}, \textit{xcomp}\}}{
        $\mathcal{L} \leftarrow$ verbs\_with\_objs$(c, \mathcal{L})$\;
    }
}
\Return{$\mathcal{L}$}\;
\caption{verbs\_with\_objs$(v, \mathcal{L})$}
\label{alg:verbs_with_obj}
\end{algorithm}

\subsubsection{Official Source \space / \space Unofficial Source.} We use the same high-level method both to identify interviews from Official Sources (e.g. police), and to determine if the article includes quotations or summarizes the perspective of an Unofficial Source (a bystander or civilian other than the victim). To do so, we consider two basic and representative phrasal forms: (1) $<$\textit{SOURCE}$>$ $<$\textit{VERB}$>$ $<$\textit{CLAUSE}$>$, and (2) \textit{according to} $<$\textit{SOURCE}$>$, $<$\textit{CLAUSE}$>$. To extract Phrase Type 1, we identify tokens of entity type PERSON or part of speech PRON such that the token is an \textit{nsubj} or \textit{nsubjpass} whose head lemma belongs to the verb set \{\textit{answer}, \textit{claim}, \textit{confirm}, \textit{declare}, \textit{explain}, \textit{reply}, \textit{report}, \textit{say}, \textit{state}, \textit{tell}\}. To extract Phrase Type 2, we identify tokens in a dependency relation\footnote{In spaCy 2.1, we need to consider two-hop relations: \\ \textit{According} (\texttt{prep}) $\rightarrow$ \textit{to} \space (\texttt{pobj}) \space $\rightarrow$ \space $<\textit{$SOURCE$}>$} with the word \textit{according}. If such a token is found in either case and it is \textit{outside} the \textsc{victim} token set, then we return that token's index as an \textbf{Unofficial Source} match. If the token is found in the \textsc{officer} set or has a lemma in \{\textit{authority}, \textit{investigator}, \textit{official}, \textit{source}\}, then we return the token's index as an \textbf{Official Source} match.

\subsubsection{Systemic claims.} This category is arguably the most variable, and as a result, possibly the most difficult to identify reliably. Systemic claims are used to frame police shootings as a product of structural or institutional racism. To identify this frame, we look for sentences that (1) mention \textit{other} police shooting incidents or (2) use certain keywords related to the national or global scope of the problem. We decide (2) using \texttt{(nation(?:[ -])?wide|wide(?:[ -])?spread|police~violence|police shootings|police~killings|racism|
racial|systemic|reform|no(?:[ -])?knock)} as our regular expression. Here, \textit{nation-wide} and \textit{widespread} indicate scope, \textit{police violence, police shootings,} and \textit{police killings} describe the persistent issue, while \textit{racism, racial, systemic} indicate the root of the issue, and \textit{reform} the solution. We also include \textit{no-knock} since there have been over 20k no-knock raids per year since the start of our data collection, and the failures of this policy have been used heavily as evidence in support of police reform \cite{lind2015cops}.  To identify (1), we match tokens $t_k$ of entity type PERSON with thematic relation PATIENT (a dependency relation in \{\textit{nsubjpass}, \textit{dobj}, \textit{iobj}, \textit{obj}\}) such that $t_k \not \in \textsc{victim}$ and $t_k \not \in \textsc{officer}$ and $t_k$ is not the object of a \textsc{victim} \textit{nsubj}. If lemma$(t_k)$ belongs to the set \{\textit{kill}, \textit{murder}, \textit{shoot}\}, we return the index of $t_k$ as a match for systemic framing. 
This process is detailed in Algorithm~\ref{alg:systemic}, with a helper function in Algorithm~\ref{alg:has_victim_subj}.

\begin{algorithm}[]
\caption{systemic$(\mathcal{D})$}
\SetAlgoLined
\KwIn{Dependency parsed document $\mathcal{D}$}
\KwOut{Document string index $i$ of the token used to identify locus of systemic framing}
\textsc{shooting} $\leftarrow$ \{\textit{shoot}, \textit{kill}, \textit{murder}\} \;
\textsc{officer}, \textsc{victim} $\leftarrow$ \text{partition}$(\mathcal{D})$ \;
\For{$t_j \in \mathcal{D}$}{
    \If{$($dep$(t_j) \in $ \{\textit{nsubjpass}, \textit{dobj}, \textit{iobj}, \textit{obj}\}$)$ \textbf{and} $($lemma$($head$(t_j)) \in$ \{\textit{shoot}, \textit{kill}, \textit{murder}\}$)$ \textbf{and} $(t_j \not \in \textsc{victim})$ \textbf{and} $(t_j \not \in \textsc{officer})$ \textbf{and} $($ent\_type$(t_j)$ = PERSON$)$ \textbf{and not} has\_victim\_subject$(t_j)$}{
        \Return{index$($head$(t_j), \mathcal{D})$}\;
    }
}
\Return{\texttt{inf}}\;
\label{alg:systemic}
\end{algorithm}

\begin{algorithm}[]
\caption{has\_victim\_subject$(o)$}
\SetAlgoLined
\KwIn{Object token $o$ from dependency parsed document}
\KwOut{\texttt{boolean}}
\For{$c \in \text{children}(\text{head}(o))$}{
    \If{$c \in \textsc{victim}$ \textbf{and} dep$(c)$ = \textit{nsubj}}{
        \Return{\texttt{true}}\;
    }
}
\Return{\texttt{false}}\;
\label{alg:has_victim_subj}
\end{algorithm}

% \subsubsection{Moral foundations.} 

% For category $\mathcal{C}$, we compute the frame centroid as
% \begin{align*}
%     \mu_{\mathcal{C}} := \frac{\sum_{w_i \in \mathcal{C}} e(w_i)} {\left\|\sum_{w_i \in \mathcal{C}} e(w_i)\right\|_2}
% \end{align*}
% for embeddings $e(w_i)$ of words $w_i \in \mathcal{C}$.

% We measure the extent to which a news document $D$ frames the officer (or alternatively, the victim) in terms of moral category $\mathcal{C}$ by taking
% \begin{equation*}
%     \texttt{score}_\text{officer}(D, \mathcal{C}) = \frac{\sum_{t_j \in M} \text{sim}(e(t_j),\mu_{\mathcal{C}})}{|M|} 
% \end{equation*}
% where
% \begin{align*}
%     \text{sim}{(x,y)} = \frac{x\cdot y}{ \|x\|\|y\| }
% \end{align*}
% is the cosine similarity, and $M \subset D$ is the set of tokens $t_j$ such that $t_j$ has a dependency relation \{\textit{amod, nn, acl}\} to some head in the \textsc{officer} (alternatively, \textsc{victim}) set, or $t_j$ is the head of some \textit{nsubj} dependant in the \textsc{officer} (\textsc{victim}) cluster. That is to say $M$ 
% contains all modifiers of officer tokens (or victim tokens) and active verbs for which an officer (or victim) is the agent.

\section{Validating Frame Extraction Methods}
\label{appdx:validation}

We report the accuracy, precision, and recall of our system in Table~\ref{tab:extraction_performance}. Ground truth is the binary presence of the frame in the 50 annotated articles above. We observe high precision and recall scores generally above 70\%, with only \textit{race} and \textit{unofficial sources} at 66\% and 65\% precision.

\begin{table}[]
    \centering
    \small
    \resizebox{\columnwidth}{!}{
    \begin{tabular}{lrrr}
         \toprule
         \textbf{Frame} & \textbf{Acc} & \textbf{Prec} & \textbf{Recall} \\ \midrule
         Age & 86\% & 100\% & 84\% \\
         Armed & 76\% & 76\% & 76\% \\
         Attack & 72\% & 73\% & 73\% \\
         Criminal record & 84\% & 77\% & 96\% \\
         Fleeing & 96\% & 89\% & 100\% \\
         Gender & 88\% & 91\% & 91\% \\
         Legal language & 88\% & 86\% & 100\% \\
         Mental illness & 100\% & 100\% & 100\% \\
         Official sources & 92\% & 95\% & 95\% \\
         Race & 92\% & 66\% & 100\% \\
         Systemic & 88\% & 86\% & 100\% \\
         Unarmed & 96\% & 88\% & 88\% \\
         Unofficial sources & 70\% & 65\% & 88\% \\
         Video & 90\% & 93\% & 78\% \\
         \bottomrule
    \end{tabular}
    }
    \caption{Frame extraction performance on 50 hand-labeled news articles}
    \label{tab:extraction_performance}
\end{table}

\section{Framing vs. Agenda Setting}
\label{appdx:supp_framing_analysis}

\begin{table}[]
    \centering
    \small
    \resizebox{\columnwidth}{!}{
    \begin{tabular}{lrrrr}
         \toprule
         \textbf{Victim Variable} & \textbf{Lib.} & \textbf{Cons.} & \textbf{Cohen's \textsl{d}}\\ \midrule 
         Mental illness  & 0.206 & 0.194 & -0.030 \\
         Fleeing  & 0.246 & 0.235 & -0.027 \\
         Video  & 0.136 & 0.155 & 0.054 \\
         Armed ** & 0.580 & 0.648 & 0.140 \\
         Attack *** & 0.351 & 0.486 & 0.276 \\
         %Unarmed ** & 0.339 & 0.279 & -0.130 \\
         \bottomrule \\
    \end{tabular}
    }
    \caption{\textbf{Agenda setting}. Proportion of liberal and conservative articles that report on killings where \textit{Victim Variable} is true (e.g. the victim really was Fleeing). We see that conservative sources report more on cases where the victim is armed and attacking}
    \label{tab:agenda_setting}
\end{table}

One potential confound is \textsl{agenda setting}, or ideological differences in the amount of coverage that is devoted to different events \cite{mccombs2002agenda}. In Table~\ref{tab:agenda_setting}, we see that conservative sources were significantly more likely to cover cases in which the victim was \textsl{armed} (.648 vs. .580, $+12\%$) and \textsl{attacking} (.486 vs. .351, $+38\%$) overall. However, we find that the differences in partisan frame alignment are \textbf{magnified} when we consider only news sources where the ground truth metadata reflects the framing category. That is, we observed larger effect sizes (Cohen's $d$) in Table~\ref{tab:conditional_partisan_framing} than we did for the observed differences in Table~\ref{tab:partisan_framing}. Furthermore, when conditioning on ground truth \textsl{race}, these frames are universally more prevalent when the victim is Black as opposed to when the victim is white. News reports on white victims thus appear more episodic \cite{lawrence2000politics}, while reports on Black victims appear to be more polarizing in terms of the given framing devices. Policing continues to be a highly racialized issue \cite{muhammad2019condemnation}.

% In Tables~\ref{tab:conditional_partisan_framing},~\ref{tab:partisan_framing_ordering},~and~\ref{tab:moral_framing}, we detail the results of our Conditional Frequency (Section~\ref{subsec:conditional_freq}), Frame Ordering (Section~\ref{subsec:frame_ordering}), and Moral Framing (Section~\ref{subsec:moral_framing_differences}) analyses.

\begin{table}[t]
    \centering
    \small
    \resizebox{\columnwidth}{!}{
    \begin{tabular}{lrrrr}
         \toprule
         \textbf{Framing Device} & \textbf{Lib.} & \textbf{Cons.} & \textbf{Cohen's \textsl{d}}\\ \midrule 
        % Age ($\leq 21$)  & 0.642 & \textbf{0.715}  & 0.156 \\ 
        % Age ($>21$) *** & 0.355 & \textbf{0.525} & 0.348 \\
        Armed (T) *** & 0.590 & \textbf{0.701} & 0.233 \\ 
        Armed (T, black) ** & 0.639 & \textbf{0.762} & 0.270 \\
        Armed (T, white)  ** & 0.552 & \textbf{0.693} & 0.293 \\ \midrule
        Attack (T) *** & 0.381 & \textbf{0.575} & 0.395 \\
        Attack (T, black) *** & 0.407 & \textbf{0.585} & 0.359 \\
        Attack (T, white) *** & 0.378 & \textbf{0.573} & 0.396 \\ \midrule
        Fleeing (T) *** & 0.381 & \textbf{0.604} & 0.458 \\
        Fleeing (T, black) * & 0.424 & \textbf{0.589} & 0.334 \\
        Fleeing (T, white) *** & 0.250 & \textbf{0.542} & 0.618 \\ \midrule
        % Gender (female victim) * & 0.557 & \textbf{0.720} & 0.340 \\ 
        % Gender (male victim)  & 0.506 & \textbf{0.524} & 0.036 \\
        Mental illness (T) * & \textbf{0.433} & 0.320 & 0.235 \\
        Mental illness (T, black) * & \textbf{0.480} & 0.291 & 0.387 \\
        Mental illness (T, white)  & \textbf{0.430} & 0.347 & 0.171 \\ \midrule
        Race (black) *** & \textbf{0.612} & 0.373 & 0.492 \\ 
        Race (white) & \textbf{0.197} & 0.146 & 0.139 \\ \midrule
        Unarmed (T) *** & \textbf{0.365} & 0.218 & 0.324 \\
        Unarmed (T, black)  & \textbf{0.441} & 0.337 & 0.212 \\
        Unarmed (T, white) ** & \textbf{0.261} & 0.118 & 0.380 \\ \midrule
        Video (T) * & 0.486 & \textbf{0.626} & 0.282 \\
        Video (T, black) & 0.529 & \textbf{0.639} & 0.223 \\
        Video (T,white) & 0.394 & \textbf{0.577} & 0.368 \\
         \bottomrule \\
    \end{tabular}
    }
    \caption{\textbf{Frame alignment is magnified when conditioned on ground truth}. The proportion of liberal and conservative news articles that include framing device conditioned on articles where ground truth reflects the framing category (T) and the victim's race is given (black, white).}
    %Significance is given by Mann-Whitney rank test:  * ($p<0.05$), ** ($p<0.01$), *** ($p<0.001$)
    \label{tab:conditional_partisan_framing}
\end{table}